\documentclass[9pt]{article}
\usepackage{spconf,amsmath,graphicx}

\usepackage{cleveref}
\usepackage{helvet}
\usepackage{tabularx}
\usepackage{enumitem}

\usepackage{algorithm}
\usepackage{algpseudocode}
\usepackage{subcaption,siunitx,booktabs}
\usepackage{amsthm}
\makeatletter
\def\thm@space@setup{%
  \thm@preskip=1pt
  \thm@postskip=\thm@preskip 
}
\makeatother

\usepackage{amssymb}

\usepackage{balance,setspace}
\usepackage{multirow,comment, color}
\usepackage{amsmath}
\usepackage{soul}
\usepackage{arydshln}
\ninept
\newlength\myindent
\def\Width{0\kern2\tabcolsep\ldots\kern1\tabcolsep0}
\usepackage{etoolbox}
\newcommand{\zerodisplayskips}{%
  \setlength{\abovedisplayskip}{3pt}
  \setlength{\belowdisplayskip}{3pt}
  \setlength{\abovedisplayshortskip}{3pt}
  \setlength{\belowdisplayshortskip}{3pt}}
\appto{\normalsize}{\zerodisplayskips}
\appto{\small}{\zerodisplayskips}
\appto{\footnotesize}{\zerodisplayskips}

\definecolor{blue}{rgb}{0, 0, 0}
\title{Low-Rank and Sparse Soft Targets To Learn Better DNN Acoustic Models}
\name{Pranay Dighe$^{\star \circ}$ \qquad Afsaneh Asaei$^{\star}$ \qquad Herv\'{e} Bourlard$^{\star \circ}$}

\address{$^{\star}$Idiap Research Institute, Martigny, Switzerland\\
$^{\circ}$\'{E}cole Polytechnique F\'{e}d\'{e}rale de Lausanne (EPFL), Switzerland}

\begin{document}
\maketitle
\begin{abstract}
Conventional deep neural networks (DNN) for speech acoustic modeling rely on Gaussian mixture models (GMM) and hidden Markov model (HMM) to obtain binary class labels as the targets for DNN training. Subword classes in speech recognition systems correspond to context-dependent tied states or senones. The present work addresses some limitations of GMM-HMM senone alignments for DNN training. We hypothesize that the senone probabilities obtained from a DNN trained with binary labels can provide more accurate targets to learn better acoustic models. However, DNN outputs bear inaccuracies which are exhibited as high dimensional unstructured noise, whereas the informative components are structured and low-dimensional. We exploit principle component analysis (PCA) and sparse coding to characterize the senone subspaces. Enhanced probabilities obtained from low-rank and sparse reconstructions are used as soft-targets for DNN acoustic modeling, that also enables training with untranscribed data. Experiments conducted on AMI corpus shows 4.6\% relative reduction in word error rate. 
\end{abstract}
\begin{keywords}
Soft targets, Principle component analysis, Sparse coding, Automatic speech recognition, Untranscribed data. 
\end{keywords}

\vspace{-2mm}
\section{Introduction}\label{sec:intro}
\vspace{-2mm}
DNN based acoustic models have been state-of-the-art for automatic speech recognition over the past few years~\cite{hinton2012deep}. While DNN input consists of multiple frames of acoustic features, the target output is obtained from a frame level GMM-HMM forced alignment corresponding to the context dependent tied triphone states or senones~\cite{young1994tree}. This procedure results in inefficiency in DNN acoustic modeling~\cite{jaitly2014autoregressive,senior2014gmmfree}.
Unlike the conventional practice, the present work argues that the optimal DNN targets are probability distributions rather than Kronecker deltas (hard targets). Earlier studies on optimal training of a neural network for HMM decoding provides rigorous theoretical analysis that supports this idea~\cite{bourlard1994remap}. Here, we propose a DNN based data driven framework to obtain accurate probability distributions (soft targets) for improved DNN acoustic modeling. The proposed approach relies on modeling of low-dimensional senone subspaces in DNN posterior probabilities. 

Speech production is known as the result of activations of a few highly constrained articulatory mechanisms leading to generation of linguistic units (e.g. phones, senones) on low-dimensional non-linear manifolds~\cite{deng2004switching,king2007speech}. In the context of DNN acoustic modeling, low-dimensional structures are exhibited in the space of DNN senone posteriors~\cite{dighe2015sparse}. Low-rank and sparse representations are found promising to characterize senone-specific subspaces~\cite{dighe2016exploiting,luyet2016lrr}. The senone-specific structures are superimposed with high-dimensional unstructured noise. Hence, projection of DNN posteriors on their underlying low-dimensional subspaces enhances the DNN posterior accuracies. 
In this work, we propose a new application of \textit{enhanced} DNN posteriors to generate accurate soft targets for DNN acoustic modeling.

Earlier works on exploiting low-dimensionality in DNN acoustic modeling focus on exploiting low-rank and sparse representations to modify DNN architectures for small footprint implementation. In~\cite{xue2013restructuring,sainath2013low} low-rank decomposition of the neural network's weight matrices enables reduction in DNN complexity and memory footprint. Similar goals have been achieved by exploiting sparse connections~\cite{yu2012exploiting} and sparse activations~\cite{liu2015neuron} in hidden layers of DNN. 
In another line of research, soft targets based DNN training has been found effective for enabling model compression~\cite{hinton2015distilling,chan2015transferring} and knowledge transfer from an accurate complex model to a smaller network~\cite{price2016wise}. This approach relies on soft targets providing more information for DNN training than the binary hard alignments.

We propose to bring together the advantage of higher information content of soft targets with the accurate model of senone space provided by low-rank and sparse representations to train superior DNN acoustic models. Soft targets enable characterization of the senone-specific subspaces by quantifying the correlations between senone classes as well as sequential dependencies (details in Section~\ref{sec:why_soft}). This information is manifested in the form of structures visible among a large population of training data posterior probabilities. Potential of these posteriors to be used as soft targets for DNN training is reduced due to presence of unstructured noise. Therefore, to obtain reliable soft targets, we perform low-rank and sparse reconstruction of training data posteriors to preserve the global low-dimensional structures while discarding the random high-dimensional noise. The new DNNs trained with low-rank or sparse soft targets are capable of estimating the test posteriors on a low-dimensional space which results in better ASR performance. 
We consider PCA (Section~\ref{sec:pca}) and dictionary based sparse coding (Section~\ref{sec:dlsc}) for generating low-rank and sparse representations respectively. Strength of PCA lies in capturing the linear regularities in the data~\cite{hutchinson2015sparse} whereas an over-complete dictionary used for sparse coding learns to model the non-linear space as a union of low-dimensional subspaces. Dictionary based sparse reconstruction also reduces the rank of the senone posterior space~\cite{dighe2016exploiting}.


%

\begin{figure*}[t]
\centering
\includegraphics[scale=0.076]{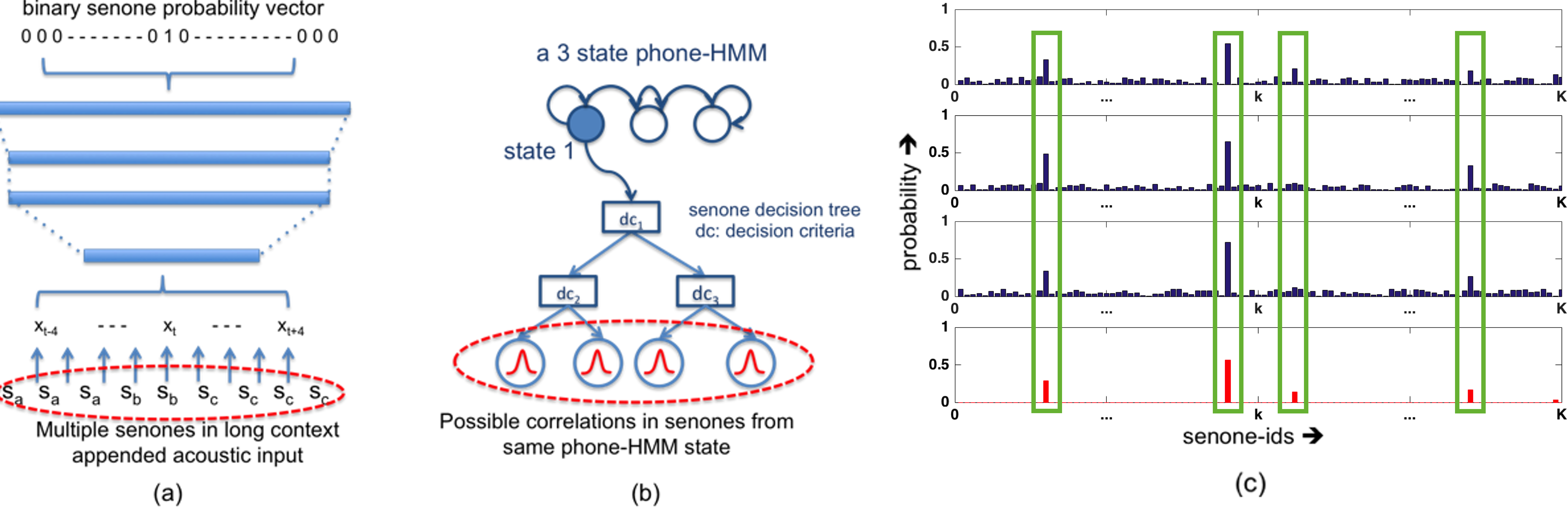} 
\vspace{-2mm}
\caption{ \footnotesize Correlation among senones due to: (a) long input context and (b) acoustically similar root in decision trees. In (c), we show examples of DNN  posterior probabilities for a particular senone class (in blue barplots) which highlight low-dimensional patterns (green boxes) super-imposed with unstructured noise. PCA and sparse coding enable recovery of the underlying patterns by discarding the unstructured noise, and provide more reliable soft targets for DNN training. $K$ denotes the size of DNN outputs which is equal to total number of senones.}\label{fig:senone-posteriors} 
\vspace{-5mm}
\end{figure*}
Experimental evaluations are conducted on AMI corpus~\cite{mccowan2005ami}, a collection  of recordings of multi-party meetings for large vocabulary speech recognition. We show in Section~\ref{sec:analysis} that low-rank and sparse soft targets lead to training of better DNN acoustic models. Reductions in word error rate (WER) are observed over the baseline hybrid DNN-HMM system without the need of explicit sparse coding or low-rank reconstruction of test data posteriors. Moreover, they enable effective use of out-of-domain untranscribed data by augmenting AMI training data in a knowledge transfer fashion. DNNs trained with low-rank and sparse soft targets yield upto 4.6\% relative improvement in WER, whereas a DNN trained with non-enhanced soft targets fails to exploit any further knowledge provided by the untranscribed data. To the best of our knowledge, significant benefit of DNN generated soft targets for training a more accurate DNN acoustic model has not been shown in the prior work. 


In the rest of the paper, the proposed approach is described in Section~\ref{sec:theory}. Experimental analysis is carried out in Section~\ref{sec:analysis}. Section \ref{sec:conclusions} presents the concluding remarks and directions for future work.

\vspace{-2mm}
\section{Low-Rank and Sparse Soft Targets}\label{sec:theory}
\vspace{-3mm}
This section describes the novel approach towards reliable soft target estimation. We study reasons for regularities among senone posteriors and investigate two systematic approaches to obtain more accurate probabilities as soft targets for DNN acoustic modeling. 
\vspace{-3mm}
\subsection{Towards Better Targets for DNN Training }\label{sec:why_soft} 
\vspace{-2mm}
Earlier works on distillation of the DNN knowledge show the potential of soft targets for model compression and the sub-optimal nature of the hard alignments~\cite{hinton2015distilling,gillick2011don}. Although hard targets assign a particular senone label to a relatively long sequence of ($\sim$10 or more) acoustic frames, senone durations are usually shorter. 
A long context of input frames may lead to presence of acoustic features corresponding to multiple senones in the input (Fig.~\ref{fig:senone-posteriors}(a)), so the assumption of binary outputs renders inaccurate. 

In contrast, soft outputs quantify such sequential information using non-zero probabilities for multiple senone classes. Contextual senone dependencies arising in soft targets can be attributed to the ambiguities due to phonetic transitions~\cite{gillick2011don}. Furthermore, the procedure of senone extraction leads to acoustic correlations among multiple classes corresponding to the same phone-HMM states~\cite{young1994tree}, as they all share the same root in the decision tree (Fig.~\ref{fig:senone-posteriors}(b)). 

These dependencies can be characterized by analyzing a large number of senone probabilities from the training data. The frequent dependencies are exhibited as regularities among the correlated dimensions in senone posteriors. 
As a result, a matrix formed by concatenation of class-specific senone posteriors has a low-rank structure. In other words, class-specific senones lie in low-dimensional subspaces with a dimension higher than unity~\cite{dighe2016exploiting}, that violates the principal assumption of binary hard targets. 

In practice, inaccuracies in DNN training lead to the presence of unstructured high-dimensional errors (Fig.~\ref{fig:senone-posteriors}(c)). 
Therefore, the initial senone posterior probabilities obtained from the forward pass of a DNN trained with hard alignments are not accurate in quantifying the senone dependency structures. Our previous work demonstrates that the erroneous estimations can be separated using low-rank or sparse representations~\cite{luyet2016lrr,dighe2016exploiting}. In the present study, we consider application of PCA and sparse coding to obtain more reliable soft targets for DNN acoustic model training.  


\begin{figure*}[t]
\centering
\includegraphics[scale=0.33]{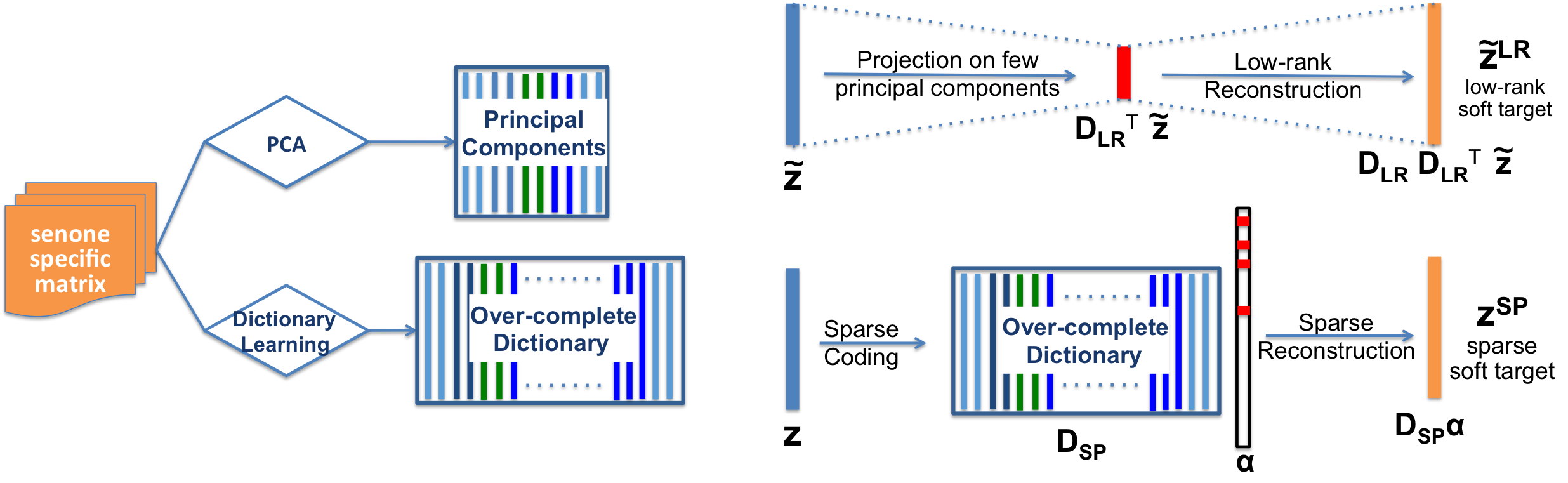} 
\vspace{-4mm}
\caption{ \footnotesize Low-dimensional reconstruction of senone posterior probabilities to achieve more accurate soft targets for DNN acoustic model training: PCA is used to extract principle components of the linear subspaces of individual senone classes. Sparse reconstruction over a dictionary of senone space representatives is used for non-linear recovery of low-dimensional structures.}\label{fig:theory}
\vspace{-6mm} 
\end{figure*}

\vspace{-3mm}
\subsection{Low-rank Reconstruction Using Eigenposteriors}\label{sec:pca}
\vspace{-2mm}
Let $z_t = [p(s_1|x_t) \hdots p(s_k|x_t) \hdots p(s_K|x_t) ]^\top$ denote a forward pass estimate of the posterior probabilities of $K$ senone classes $\{s_k\}_{k=1}^K$, given the acoustic feature $x_t$ at time $t$. DNN is trained using the initial labels obtained from GMM-HMM forced alignment. We collect $N$ senone posteriors which are labeled as class $s_k$ in GMM-HMM forced alignment and mean-center them in the \textit{logarithmic} domain as follows:
\begin{equation}\label{eq:log_post}
\tilde{z}_t = \ln(z_t) - \mu_{s_k}
\end{equation} 
where $\mu_{s_k}$ is mean of the collected posteriors in log-domain. Due to skewed distribution of the posterior vectors, the logarithm of posteriors fits better the Gaussian assumption of PCA. We concatenate the $N$ senone $s_k$ posterior vectors after operation shown in \eqref{eq:log_post} to form a matrix $M_{s_k} \in \mathcal{R}^{K \times N}$. For the sake of brevity, the subscript $s_k$ is dropped in the subsequent expressions. However, all the calculations are performed for each of the senone classes individually. 

Principal components of the senone space are obtained via eigenvector decomposition~\cite{shlens2014tutorial} of covariance matrix of $M$. 
The covariance matrix is obtained as $C = \frac{1}{N-1}M M^\top$. We factorize the covariance matrix as $C = P S P^\top$ where $P\in\mathcal{R}^{K\times K}$ identifies the eigenvectors and $S$ is a diagonal matrix containing the sorted eigenvalues.  
Eigenvectors in $P$ which correspond to the large eigenvalues in $S$ constitute the frequent regularities in the subspace, whereas others carry the high-dimensional unstructured noise. 
Hence, the low-rank projection matrix is defined as 
\begin{equation}\label{eq:pca}
D_{\text{LR}} = P_l\in \mathcal{R}^{K\times l}
\end{equation}
where $P_l$ is truncation of $P$ that keeps only the first $l$ eigenvectors and discards the erroneous variability captured by other $K-l$ components. We select $l$ such that relatively $\sigma$\% variability is preserved in the low-rank reconstruction of original senone matrix $M$. 

The eigenvectors stored in the low-rank projection $P_l$ are referred to as \textit{``eigenposteriors''} of the senone space (in the same spirit as \textit{eigenfaces} are defined for low-dimensional modeling of human faces~\cite{sirovich87faces}). 

Low-rank reconstruction of a mean-centered log posterior $\tilde{z}_t$, denoted by $\tilde{z}_t^{\text{LR}}$ is estimated as 
\begin{equation}
\tilde{z}_t^{\text{LR}} = D_{\text{LR}} {D_{\text{LR}}}^\top  \tilde{z}_t  
\end{equation}

Finally, we add the mean $\mu_{s_k}$ to $\tilde{z}_t^{\text{LR}}$ and take its exponent to obtain a low-rank senone posterior $z^{LR}_t$ for the acoustic frame $x_t$. Low-rank posteriors obtained for the training data are used as soft targets for learning better DNNs (Fig.\ref{fig:theory}).
We assume that $\sigma$\% variability, that quantifies the low-rank regularities in senone spaces, is a parameter independent of the senone class.

\vspace{-3mm}
\subsection{Sparse Reconstruction Using Dictionary Learning}\label{sec:dlsc}
\vspace{-1mm}
Unlike PCA, over-complete dictionary learning and sparse coding enables modeling of non-linear low-dimensional manifolds. Sparse modelling assumes that senone posteriors can be generated as sparse linear combination of senone space representatives, collected in a dictionary $D_{\text{SP}}$.
We use online dictionary learning algorithm~\cite{mairal2010online} to learn an over-complete dictionary for senone $s_k$ using a collection of $N$ training data posteriors of senone $s_k$, such that 
\begin{equation}\small \label{eq:dict_learn}
D_{\text{SP}} = \arg \min_{\substack D,A} \sum_{t=t_1}^{t_N} \| z_t - D\,\alpha_{t} \|_2^2 + \lambda{\|\alpha_{t}\|}_1
\end{equation}
where $A=[\alpha_{t_1}\hdots\alpha_{t_N}]$ and $\lambda$ is a regularization factor.  Again we have dropped the subscript $s_k$, but all calculations are still senone-specific. 
Sparse reconstruction (Fig.\ref{fig:theory}) of senone posteriors is thus obtained by first estimating the sparse representation~\cite{tibshirani1996regression} as 
\begin{equation}\small \label{eq:sparse}
\alpha_{t} = \arg \min_{\alpha} \| z_t - D_{\text{SP}}\,\alpha \|_2^2 + \lambda{\|\alpha\|}_1. 
\end{equation}
followed by reconstruction as 
\begin{equation}\small \label{eq:reconst}
z^{SP}_t = D_{\text{SP}}\; \alpha_{t} \qquad \forall t \in \{t_1,\hdots,t_N\}.
\end{equation}

Sparse reconstructed senone posteriors have been previously found to be more accurate acoustic models for DNN-HMM speech recognition~\cite{dighe2016exploiting}. In particular, it was shown that the rank of senone-specific matrices is much lower after sparse reconstruction. In the present work, we investigate if they could also provide more accurate soft targets for DNN training
Regularization parameter $\lambda$ in~\eqref{eq:dict_learn}-\eqref{eq:sparse} controls the level of sparsity and the level of noise being removed after sparse reconstruction. 
Fig.~\ref{fig:theory} summarises the low-rank and sparse reconstruction of senone posteriors.
\vspace{-2.5mm}
\section{Experimental Analysis}\label{sec:analysis}
\vspace{-2mm}
In this section we evaluate the effectiveness of low-rank and sparse soft targets to improve the performance of DNN-HMM speech recognition. We also investigate the importance of better DNN acoustic models to exploit information from untranscribed data. 

\begin{table*}[t]\footnotesize
\caption{\footnotesize Performance of various systems (in WER\%) when additional untranscribed training data is used. System 0 is hard-targets based baseline DNN. In paranthesis, SE-0 denotes supervised enhancement of DNN outputs from system 0 and FP-$n$ shows forward pass using system $n$.}
\vspace{-3mm}
\centering
\begin{tabular}{c l c c || c}
\hline 
System \# &Training Data & PCA($\sigma$=80) & Sparsity($\lambda$=0.1) & Non-Enhanced Soft-Targets\\
\hline 
0 & AMI   (Baseline WER \textbf{32.4\%}) & - & - & -\\
1 & AMI(SE-0) & 31.9 & 31.6 & 32.0\\
2 & ICSI(FP-1) + AMI(SE-0) & 31.2 & 31.6 & 32.5\\
3 & LIB100(FP-1) + AMI(SE-0) & 31.2 & 31.6 & 32.4\\
4 & LIB100(FP-2) + AMI(SE-0) & 31.0 & 31.8 & 32.4\\
5 & LIB100(FP-2) + ICSI(FP-2) + AMI(SE-0) & \textbf{30.9} & 31.7 & 32.4\\
\hline 
\end{tabular} 
\vspace{-4mm}
\label{table:data_augmentation}
\end{table*}

\vspace{-3.5mm}
\subsection{Database and Speech Features}\label{sec:database}
\vspace{-2mm}
Experiments are conducted on AMI corpus~\cite{mccowan2005ami} which contains recordings of spontaneous conversations in meeting scenarios. We use recordings from individual head microphones (IHM) comprising of around 67 hours of \textit{train} set, 9 hours of development, (\textit{dev}) set, and 7 hours \textit{test} set. 10\% of training data is used for cross-validation during DNN training, whereas \textit{dev} set is used for tuning regularization parameters $\sigma$ and $\lambda$. For experiments using untranscribed additional training data, we use ICSI meeting corpus~\cite{janin2003icsi} and Librispeech corpus~\cite{panayotov2015librispeech}. Data from ICSI corpus consists of meeting recordings (around 70 hours). Librispeech data is read speech from audio-books and we use a 100 hour subset of it. 

Kaldi toolkit~\cite{povey2011kaldi} is used for training DNN-HMM systems. All DNNs have 9 frames of temporal context at acoustic input and 4 hidden layers with 1200 neurons each. Input features are 39 dimensional MFCC+$\Delta$+$\Delta\Delta$ (39$\times$9=351 dimensional input) and output is 4007 dimensional senone probability vector. AMI pronunciation dictionary has $\sim$23K words and a bigram model for decoding. For dictionary learning and sprase coding, SPAMS toolbox~\cite{mairal2014sparse} is used.

\vspace{-3.5mm}
\subsection{Baseline DNN-HMM using Hard and Soft Targets}\label{sec:results}
\vspace{-2mm}
Our baseline is a hybrid DNN-HMM system trained using forced aligned targets (IHM setup in~\cite{himawan2015learning}). WER using baseline DNN is 32.4\% on AMI \textit{test} set. Another baseline is a DNN trained using non-enhanced soft targets from the baseline. This system gives a WER of 32.0\%. All soft-target based DNNs are randomly initialized and trained using cross-entropy loss backpropagation. 

\vspace{-3.5mm}
\subsection{Generation of Low-rank and Sparse Soft Targets}\label{sec:rkt}
\vspace{-2mm}
We group DNN forward pass senone probabilities for the training data into class-specific senone matrices. For this, senone labels from the ground truth based GMM-HMM hard alignments are used. Each matrix is restricted to have $N=10^4$ vectors of $K=4007$ senone probabilities to facilitate computation of principal components and sparse dictionary learning. We found the average rank of senone matrices, defined as the number of singular values required to preserve 95\% variability, to be 44. Dictionaries of size 500 columns were learned for each senone, making them nearly 10 times over-complete. The procedure as depicted in Fig.~\ref{fig:theory} is implemented to generate low-rank and sparse soft-targets. 

We also encountered memory issues while storing large matrices of senone probabilities for all training and cross-validation data. It requires enormous amounts of storage space (similar to~\cite{chan2015transferring}). Hence, we preserve precision only upto first two decimal places in soft targets, followed by normalizing the vector to sum 1 before storing on the disk. We assume that essential information might not be in dimensions with very small probabilities. Although such thresholding can be a compromise to our approach, we did some experiments with higher precision (upto 5 decimal places), but there was no significant improvement in ASR. Both low-rank and sparse reconstruction were still computed on full soft-targets without any rounding; we perform thresholding only when storing targets on the disk.

First we tune the variability preserving low-rank reconstruction parameter $\sigma$ and sparsity regularizer $\lambda$ for better ASR performance in AMI \textit{dev} set. When $\sigma$=80\% of variability is preserved in the principal components space, the most accurate soft targets are achieved for DNN acoustic modeling resulting in the smallest WER. Likewise, $\lambda=0.1$ was found the optimal value for sparse reconstruction. It may be noted that in both low-rank and sparse reconstruction, there is an optimal amount of enhancement needed for improving ASR. While less enhancement leads to continued presence of noise in soft targets, too much of it results in loss of essential information.

%

\vspace{-3.5mm}
\subsection{DNN-HMM Speech Recognition}\label{sec:rkt}
\vspace{-2mm}
Speech recognition using DNNs trained with the new soft targets obtained from low-rank and sparse reconstruction is compared in Table~\ref{table:data_augmentation}). System-0 is the baseline hard target based DNN. System-1 is built by supervised enhancement of soft outputs obtained from system-0 on AMI training data as shown in Fig.~\ref{fig:theory}. As expected, training with the soft targets yields lower WER than the baseline hard targets. We can see that both PCA and sparse reconstruction result in more accurate acoustic modeling, where sparse reconstruction achieves 0.8\% absolute reduction in WER. 

Sparse reconstruction is found to work better than low-rank reconstruction for ASR. It can be due to the higher accuracy of sparse model in characterizing the non-linear senone subspaces~\cite{dighe2015sparse}. 
Unlike previous works~\cite{dighe2016exploiting,luyet2016lrr} which required two stages of DNN forward pass and explicit low-dimensional projection, a single DNN is learned here that estimates the probabilities directly on a low-dimensional space. 

%
\vspace{-3.5mm}
\subsection{Training with Untranscribed Data}\label{sec:untranscribed}
\vspace{-2mm}
Given an accurate DNN acoustic model and some untranscribed input speech data, we can obtain soft targets for the new data through forward pass.
Assuming that the initial model can generalize well on unseen data, the additional soft targets thus generated can be used to augment our original training data. We propose to learn better DNN acoustic models using this augmented training set. This method is reminiscent of the \textit{knowledge transfer} approach~\cite{hinton2015distilling,chan2015transferring} which is typically used for model compression. In this work, we use the same network architecture for all experiments.

DNNs trained with low-rank and sparse soft targets are used to generate soft targets for ICSI corpus and Librispeech (LIB100) which are sources of untranscribed data. Table~\ref{table:data_augmentation} shows interesting observations from various experiments using data augmentation. First, system-2 is built augmenting enhanced AMI training data with ICSI soft targets generated from system-1. We consider ICSI corpus, consisting of spontaneous speech from meeting recordings, as in-domain with AMI corpus. While PCA based DNN successfully exploits information from this additional ICSI data showing significant improvement from system-1 to system-2, the same is not observed using sparsity based DNN.

Next, system-3 is built by augmenting enhanced AMI data with Librispeech(LIB100) soft targets obtained from system 1. Read audio book speech data from Librispeech is out-of-domain as compared to spontaneous speech in AMI. Still, system-3 achieves similar reductions in WER as observed in system-2 which was built using in-domain ICSI data. 

System 4 and 5 were built to further explore if we could extract even more information from the out-of-domain Librispeech data by using soft targets from system-2 instead of system-1. Note that system-2, trained using soft targets from both AMI and ICSI spontaneous speech data, is a more accurate model than system 1. Indeed, both system 4 and 5 perform better than previous systems using PCA based DNNs where system 5 outperforms the hard target based baseline by 1.5\% absolute reduction in WER.

Surprisingly, DNN soft targets obtained from sparse reconstruction are not able to exploit the unseen data in all the systems. 
We speculate that dictionary learning for sparse coding captures the non-linearities specific to AMI database. These nonlinear characteristics may correspond to channel and recording conditions which vary over different databases and can not be transcended. On the other hand, the local linearity assumption of PCA leads to extraction of a highly restricted basis set that captures the most important dynamics in the senone probability space. Such regularities mainly address the acoustic dependencies among senones which are generalizable to other acoustic conditions. Hence, the eigenposteriors are invariant to the exceptional effects due to channel and recording conditions. 

Sparse reconstruction is able to mitigate the undesired effects as long as they have been seen in the training data. Given the superior performance of sparse reconstruction of AMI posteriors (in system-1), we believe that sparse modeling might be more powerful if some labeled data from unseen acoustic conditions is made available for dictionary learning.

It may be noted that training with additional untranscribed data is not effective if non-enhanced soft targets are used. In fact, systems 2-5 without low-rank or sparse reconstruction, perform worse than system-1 although they have seen more training data. 
 \vspace{-2mm}
\section{Conclusions and Future Directions}\label{sec:conclusions}
 \vspace{-2mm}
We presented a novel approach to improve DNN acoustic model training using low-rank and sparse soft targets. 
PCA and sparse coding were employed to identify senone subspaces, and enhance senone probabilities through low-dimensional reconstruction. Low-rank reconstruction using PCA relies on the existance of eigenposteriors capturing the local dynamics of senone subspaces. 
Although, sparse reconstruction proves more effective to achieve reliable soft targets when transcribed data is provided, low-rank reconstruction is found generalizable to out-of-domain untranscribed data.
DNN trained on low-rank reconstruction acheives 4.6\% relative reduction in WER, whereas DNN trained using non-enhanced soft targets fails to exploit additional information from additional data. Eigenposteriors can be better estimated using robust PCA~\cite{liu2013robust} and sparse PCA~\cite{zou2006sparse} for better modeling of senone subspaces. Furthermore, probabilistic PCA and maximum likelihood eigen decomposition can reduce the computational cost for large scale applications. 

This study supports the use of probabilistic outputs for DNN acoustic modeling. 
Specifically, enhanced soft targets can be more effective in training small footprint DNNs based on model compression. In future, we plan to investigate their usage in cross-lingual knowledge transfer~\cite{swietojanski2012unsupervised}. We will also study domain adaptation based on the notion of eigenposteriors.



 
\vspace{-3mm}
\section{Acknowledgments}
\vspace{-2mm}
{ Research leading to these results has received funding from SNSF project on ``Parsimonious Hierarchical Automatic Speech Recognition (PHASER)'' grant agreement number 200021-153507.}

\newpage
\small
\bibliographystyle{IEEEbib}
\bibliography{MainFile}

\begin{thebibliography}{10}

\bibitem{hinton2012deep}
Geoffrey Hinton, Li~Deng, Dong Yu, George~E Dahl, Abdel-rahman Mohamed, Navdeep
  Jaitly, Andrew Senior, Vincent Vanhoucke, Patrick Nguyen, Tara~N Sainath,
  et~al.,
\newblock ``Deep neural networks for acoustic modeling in speech recognition:
  The shared views of four research groups,''
\newblock {\em IEEE Signal Processing Magazine}, vol. 29, no. 6, pp. 82--97,
  2012.

\bibitem{young1994tree}
Steve~J Young, Julian~J Odell, and Philip~C Woodland,
\newblock ``Tree-based state tying for high accuracy acoustic modelling,''
\newblock in {\em Proceedings of the workshop on Human Language Technology}.
  Association for Computational Linguistics, 1994.

\bibitem{jaitly2014autoregressive}
Navdeep Jaitly, Vincent Vanhoucke, and Geoffrey Hinton,
\newblock ``Autoregressive product of multi-frame predictions can improve the
  accuracy of hybrid models,''
\newblock 2014.

\bibitem{senior2014gmmfree}
A.~Senior, G.~Heigold, M.~Bacchiani, and H.~Liao,
\newblock ``Gmm-free dnn acoustic model training,''
\newblock in {\em IEEE ICASSP}, 2014.

\bibitem{bourlard1994remap}
Herve Bourlard, Yochai Konig, and Nelson Morgan,
\newblock {\em {REMAP}: Recursive Estimation and Maximization of a Posteriori
  Probabilities; Application to Transition-based Connectionist Speech
  Recognition},
\newblock ICSI Technical Report TR-94-064, 1994.

\bibitem{deng2004switching}
Li~Deng,
\newblock ``Switching dynamic system models for speech articulation and
  acoustics,''
\newblock in {\em Mathematical Foundations of Speech and Language Processing},
  pp. 115--133. Springer New York, 2004.

\bibitem{king2007speech}
Simon King, Joe Frankel, Karen Livescu, Erik McDermott, Korin Richmond, and
  Mirjam Wester,
\newblock ``Speech production knowledge in automatic speech recognition,''
\newblock {\em The Journal of the Acoustical Society of America}, 2007.

\bibitem{dighe2015sparse}
Pranay Dighe, Afsaneh Asaei, and Herv{\'e} Bourlard,
\newblock ``Sparse modeling of neural network posterior probabilities for
  exemplar-based speech recognition,''
\newblock {\em Speech Communication}, 2015.

\bibitem{dighe2016exploiting}
Pranay Dighe, Gil Luyet, Afsaneh Asaei, and Herve Bourlard,
\newblock ``Exploiting low-dimensional structures to enhance dnn based acoustic
  modeling in speech recognition,''
\newblock in {\em IEEE ICASSP}, 2016.

\bibitem{luyet2016lrr}
Gil Luyet, Pranay Dighe, Afsaneh Asaei, and Herv{\'{e}} Bourlard,
\newblock ``Low-rank representation of nearest neighbor phone posterior
  probabilities to enhance dnn acoustic modeling,''
\newblock in {\em Interspeech}, 2016.

\bibitem{xue2013restructuring}
Jian Xue, Jinyu Li, and Yifan Gong,
\newblock ``Restructuring of deep neural network acoustic models with singular
  value decomposition.,''
\newblock in {\em INTERSPEECH}, 2013.

\bibitem{sainath2013low}
Tara~N Sainath, Brian Kingsbury, Vikas Sindhwani, Ebru Arisoy, and Bhuvana
  Ramabhadran,
\newblock ``Low-rank matrix factorization for deep neural network training with
  high-dimensional output targets,''
\newblock in {\em IEEE ICASSP}, 2013.

\bibitem{yu2012exploiting}
Dong Yu, Frank Seide, Gang Li, and Li~Deng,
\newblock ``Exploiting sparseness in deep neural networks for large vocabulary
  speech recognition,''
\newblock in {\em IEEE ICASSP}, 2012.

\bibitem{liu2015neuron}
Jian Kang, Cheng Lu, Meng Cai, Wei-Qiang Zhang, and Jia Liu,
\newblock ``Neuron sparseness versus connection sparseness in deep neural
  network for large vocabulary speech recognition,''
\newblock in {\em ICASSP}, April 2015, pp. 4954--4958.

\bibitem{hinton2015distilling}
Geoffrey Hinton, Oriol Vinyals, and Jeff Dean,
\newblock ``Distilling the knowledge in a neural network,''
\newblock {\em arXiv preprint arXiv:1503.02531}, 2015.

\bibitem{chan2015transferring}
William Chan, Nan~Rosemary Ke, and Ian Lane,
\newblock ``Transferring knowledge from a rnn to a dnn,''
\newblock in {\em Interspeech}, 2015.

\bibitem{price2016wise}
Ryan Price, Ken-ichi Iso, and Koichi Shinoda,
\newblock ``Wise teachers train better dnn acoustic models,''
\newblock {\em EURASIP Journal on Audio, Speech, and Music Processing}, , no.
  1, pp. 1--19, 2016.

\bibitem{hutchinson2015sparse}
Brian Hutchinson, Mari Ostendorf, and Maryam Fazel,
\newblock ``A sparse plus low-rank exponential language model for limited
  resource scenarios,''
\newblock {\em IEEE Transactions on Audio, Speech, and Language Processing},
  vol. 23, no. 3, pp. 494--504, 2015.

\bibitem{mccowan2005ami}
Iain McCowan, Jean Carletta, W~Kraaij, S~Ashby, S~Bourban, M~Flynn,
  M~Guillemot, T~Hain, J~Kadlec, V~Karaiskos, et~al.,
\newblock ``The ami meeting corpus,''
\newblock in {\em Proceedings of the 5th International Conference on Methods
  and Techniques in Behavioral Research}, 2005, vol.~88.

\bibitem{gillick2011don}
Dan Gillick, Larry Gillick, and Steven Wegmann,
\newblock ``Don't multiply lightly: Quantifying problems with the acoustic
  model assumptions in speech recognition,''
\newblock in {\em IEEE Workshop on Automatic Speech Recognition and
  Understanding (ASRU)}, 2011.

\bibitem{shlens2014tutorial}
Jonathon Shlens,
\newblock ``A tutorial on principal component analysis,''
\newblock {\em arXiv preprint arXiv:1404.1100}, 2014.

\bibitem{sirovich87faces}
L.~Sirovich and M.~Kirby,
\newblock ``Low-dimensional procedure for the characterization of human
  faces,''
\newblock {\em J. Opt. Soc. Am. A}, pp. 519--524, 1987.

\bibitem{mairal2010online}
Julien Mairal, Francis Bach, Jean Ponce, and Guillermo Sapiro,
\newblock ``Online learning for matrix factorization and sparse coding,''
\newblock {\em Journal of Machine Learning Research (JMLR)}, vol. 11, pp.
  19--60, 2010.

\bibitem{tibshirani1996regression}
Robert Tibshirani,
\newblock ``Regression shrinkage and selection via the lasso,''
\newblock {\em Journal of the Royal Statistical Society. Series B
  (Methodological)}, pp. 267--288, 1996.

\bibitem{janin2003icsi}
Adam Janin, Don Baron, Jane Edwards, Dan Ellis, David Gelbart, Nelson Morgan,
  Barbara Peskin, Thilo Pfau, Elizabeth Shriberg, Andreas Stolcke, et~al.,
\newblock ``The icsi meeting corpus,''
\newblock in {\em IEEE ICASSP}, 2003.

\bibitem{panayotov2015librispeech}
Vassil Panayotov, Guoguo Chen, Daniel Povey, and Sanjeev Khudanpur,
\newblock ``Librispeech: an asr corpus based on public domain audio books,''
\newblock in {\em IEEE ICASSP}, 2015.

\bibitem{povey2011kaldi}
Daniel Povey, Arnab Ghoshal, Gilles Boulianne, Luk{\'a}{\v{s}} Burget,
  Ond{\v{r}}ej Glembek, Nagendra Goel, Mirko Hannemann, Petr
  Motl{\'\i}{\v{c}}ek, Yanmin Qian, Petr Schwarz, et~al.,
\newblock ``The kaldi speech recognition toolkit,''
\newblock 2011.

\bibitem{mairal2014sparse}
Julien Mairal, Francis Bach, and Jean Ponce,
\newblock ``Sparse modeling for image and vision processing,''
\newblock {\em arXiv preprint arXiv:1411.3230}, 2014.

\bibitem{himawan2015learning}
I.~Himawan, P.~Motlicek, D.~Imseng, B.~Potard, N.~Kim, and J.~Lee,
\newblock ``Learning feature mapping using deep neural network bottleneck
  features for distant large vocabulary speech recognition,''
\newblock in {\em IEEE ICASSP}, 2015, pp. 4540--4544.

\bibitem{liu2013robust}
Guangcan Liu, Zhouchen Lin, Shuicheng Yan, Ju~Sun, Yong Yu, and Yi~Ma,
\newblock ``Robust recovery of subspace structures by low-rank
  representation,''
\newblock {\em IEEE Transactions on Pattern Analysis and Machine Intelligence},
  , no. 99, pp. 1--1, 2013.

\bibitem{zou2006sparse}
Hui Zou, Trevor Hastie, and Robert Tibshirani,
\newblock ``Sparse principal component analysis,''
\newblock {\em Journal of computational and graphical statistics}, vol. 15, no.
  2, pp. 265--286, 2006.

\bibitem{swietojanski2012unsupervised}
Pawel Swietojanski, Arnab Ghoshal, and Steve Renals,
\newblock ``Unsupervised cross-lingual knowledge transfer in dnn-based lvcsr,''
\newblock in {\em IEEE Spoken Language Technology Workshop (SLT)}, 2012.

\end{thebibliography}

\end{document}